\documentclass[acmlarge]{acmart}

\usepackage[ruled,vlined,linesnumbered]{algorithm2e}
\usepackage{booktabs} 
\usepackage{siunitx}
\usepackage{multirow}
\usepackage{subfigure}
\usepackage{enumerate}
\usepackage{amsmath}
\usepackage{makecell}

\SetAlFnt{\small}
\SetAlCapFnt{\small}
\SetAlCapNameFnt{\small}
\SetAlCapHSkip{0pt}
\IncMargin{-\parindent}

\DeclareMathOperator*{\argmin}{argmin}

\setcopyright{acmcopyright}
\acmJournal{IMWUT}
\acmYear{2019} 
\acmVolume{0} 
\acmNumber{0} 
\acmArticle{0}
\acmMonth{0} 
\acmPrice{15.00}
\acmDOI{12.1234/1234567}

\received{November 2019}

\begin{document}

\title{VLUC: An Empirical Benchmark for Video-Like Urban Computing on Citywide Crowd and Traffic Prediction} 


\author{Renhe Jiang*}
\thanks{*Equal contribution}
\affiliation{\institution{The University of Tokyo}}
\email{jiangrh@csis.u-tokyo.ac.jp}

\author{Zekun Cai*}
\affiliation{\institution{The University of Tokyo}}
\email{caizekun@csis.u-tokyo.ac.jp}

\author{Zhaonan Wang}
\affiliation{\institution{The University of Tokyo}}
\email{znwang@csis.u-tokyo.ac.jp}

\author{Chuang Yang}
\affiliation{\institution{The University of Tokyo}}
\email{chuang.yang@csis.u-tokyo.ac.jp}

\author{Zipei Fan}
\affiliation{\institution{The University of Tokyo}}
\email{fanzipei@iis.u-tokyo.ac.jp}

\author{Xuan Song}
\affiliation{\institution{SUSTech-UTokyo Joint Research Center on Super Smart City, Department of Computer Science and Engineering, Southern University of Science and Technology (SUSTech)}}
\email{songxuan@csis.u-tokyo.ac.jp}

\author{Kota Tsubouchi}
\affiliation{\institution{Yahoo Japan Corporation}}
\email{ktsubouc@yahoo-corp.jp}

\author{Ryosuke Shibasaki}
\affiliation{\institution{The University of Tokyo}}
\email{shiba@csis.u-tokyo.ac.jp}

\begin{abstract}
	Nowadays, massive urban human mobility data are being generated from mobile phones, car navigation systems, and traffic sensors. Predicting the density and flow of the crowd or traffic at a citywide level becomes possible by using the big data and cutting-edge AI technologies. It has been a very significant research topic with high social impact, which can be widely applied to emergency management, traffic regulation, and urban planning. In particular, by meshing a large urban area to a number of fine-grained mesh-grids, citywide crowd and traffic information in a continuous time period can be represented like a video, where each timestamp can be seen as one video frame. Based on this idea, a series of methods have been proposed to address video-like prediction for citywide crowd and traffic. In this study, we publish a new aggregated human mobility dataset generated from a real-world smartphone application and build a standard benchmark for such kind of video-like urban computing with this new dataset and the existing open datasets. We first comprehensively review the state-of-the-art works of literature and formulate the density and in-out flow prediction problem, then conduct a thorough performance assessment for those methods. With this benchmark, we hope researchers can easily follow up and quickly launch a new solution on this topic.
\end{abstract}


\maketitle

\section{Introduction}

\begin{figure}[t]
	\centering	
	\includegraphics[width=0.98\textwidth]{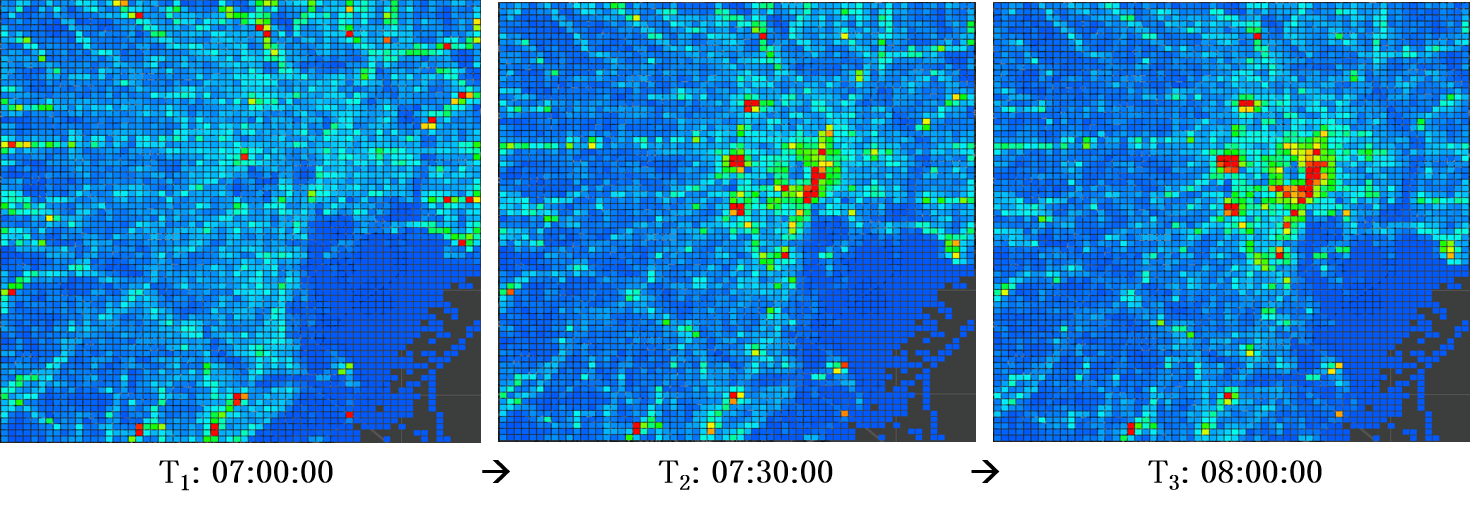}
	\caption{An illustration of citywide crowd density in Tokyo from 07:00:00 to 08:00:00 is shown above, where the red color represents higher density and the blue color represents lower density. By meshing a large urban area to a number of fine-grained mesh-grids, citywide crowd density in a continuous time period can be analogously seen as a short video.}
	\label{fig:intro}
\end{figure}

Nowadays, massive urban human mobility data are being generated from mobile phones, car navigation systems, and traffic sensors. Many studies have analyzed these big mobility data with cutting-edge technologies, which have been summarized as urban computing by \cite{zheng2014urban}. In particular, crowd or traffic prediction at a citywide level becomes an emerging topic in both academia and industry, as it can be of great importance for emergency management, traffic regulation, and urban planning. As illustrated in Fig.\ref{fig:intro}, by meshing a large urban area to a number of fine-grained mesh-grids, citywide crowd and traffic information in a continuous time period can be represented with a four-dimensional tensor $\mathbb{R}^{Timestep\times Height\times Width\times Channel}$ in an analogous manner to video data, where each $Timestep$ can be seen as one video frame, $Height$, $Width$ is two-dimensional index for mesh-grids, and each $Channel$ stores an aggregated scalar value for each mesh-grid. Following this representation, as shown in Table \ref{tab:modelsummary}, a series of studies \cite{hoang2016forecasting,zhang2016dnn,zhang2016xiuwen,zhang2017deep,wang2017deepsd,yao2018deep,zonoozi2018periodic,zhang2018predicting,yuan2018hetero,zhang2019flow,yao2019revisiting,lin2019deepstn+,jiang2019deepurbanevent} have been conducted to address video-like urban computing problems such as crowd in-out flow prediction, taxi demand prediction, and traffic accident prediction. These forecasts can be provided to governments (e.g. police) and public service operators (e.g. subway or bus company) to protect people's safety or maintain the operation of public infrastructures under event situation (e.g. New Year Countdown); to ride-sharing companies like Uber and Didi Chuxing to more effectively dispatch the taxis; to web mapping services like Yahoo Map\footnote{https://map.yahoo.co.jp/maps?layer=crowd\&v=3} and Itsumo-Navi\footnote{https://lab.its-mo.com/densitymap/} to improve the functionality of crowd density map service. 

\begin{table*}[h]%
	\centering
	\caption{Summary of The State-Of-The-Arts}
	\label{tab:modelsummary}
	\begin{tabular*}{15.7cm}{@{\extracolsep{\fill}}|l|l|l|l|}
		\hline
		Model & Reference & Dataset (* means Open) & Prediction Task\\
		\hline
		ST-ResNet\cite{zhang2017deep} & AAAI17 &  TaxiBJ*, BikeNYC* & Taxi In-Out Flow (Traffic)\\
		DeepSD\cite{wang2017deepsd}  & ICDE17& Didi Taxi Request & Taxi Demand (Traffic)\\
		DMVST-Net\cite{yao2018deep}  & AAAI18&  Didi Taxi Request & Taxi Demand (Traffic)\\
		Periodic-CRN\cite{zonoozi2018periodic} & IJCAI18 & TaxiBJ*, TaxiSG & Taxi Density/In-Out Flow (Traffic)\\
		Hetero-ConvLSTM\cite{yuan2018hetero}  & KDD18& Vehicle Crash Data* & Traffic Accident (Traffic)\\
		STDN\cite{yao2019revisiting} & AAAI19 & TaxiNYC*, BikeNYC-II* & Taxi/Bike O-D Number (Traffic)\\
		DeepSTN+\cite{lin2019deepstn+}  & AAAI19 & MobileBJ, BikeNYC-I* & Crowd/Taxi In-Out Flow (Crowd\&Traffic)\\
		Multitask Model\cite{zhang2019flow} & TKDE19 & TaxiBJ, BikeNYC & Taxi/Bike In-Out Flow (Traffic)\\
		DeepUrbanEvent\cite{jiang2019deepurbanevent} & KDD19 & Konzatsu Toukei & Crowd Density/Flow (Crowd)\\
		\hline
	\end{tabular*}
\end{table*}%

\begin{table*}[h]%
	\centering
	\caption{Details of The State-Of-The-Arts}
	\label{tab:modeldetail}
	\begin{tabular*}{15cm}{@{\extracolsep{\fill}}|l|l|l|l|l|}
		\hline
		Model & Comparison & Objective Function & Metric & Extra Data\\
		\hline
		ST-ResNet\cite{zhang2017deep} & \cite{zhang2016dnn} & MSE & RMSE & Weather, Event\\
		DeepSD & NA & MSE & MAE, RMSE & Weather, Traffic\\
		DMVST-Net\cite{yao2018deep} & \cite{zhang2017deep} & Self-Defined & MAPE, RMSE & Weather, Date\\
		Periodic-CRN\cite{zonoozi2018periodic} & \cite{zhang2017deep} & MSE & RMSE & Date\\
		Hetero-ConvLSTM\cite{yuan2018hetero} & NA & Cross Entropy (CE) & MSE, RMSE, CE & Weather, Road, etc.\\
		STDN\cite{yao2019revisiting} & \cite{wang2017deepsd}\cite{zhang2017deep}\cite{yao2018deep} & MSE & RMSE, MAPE & Weather, Event\\
		DeepSTN+\cite{lin2019deepstn+} & \cite{zhang2017deep} & RMSE & RMSE, MAE & Date, PoI\\
		Multitask Model\cite{zhang2019flow} & \cite{zhang2016dnn}\cite{zhang2017deep} & Self-Defined & RMSE, MAE & Weather, Event\\
		DeepUrbanEvent\cite{jiang2019deepurbanevent} & \cite{zhang2017deep}\cite{fan2015citymomentum} & MSE & MSE & NA\\
		\hline
	\end{tabular*}
\end{table*}%

\begin{figure}[h]
	\centering	
	\includegraphics[width=0.99\textwidth]{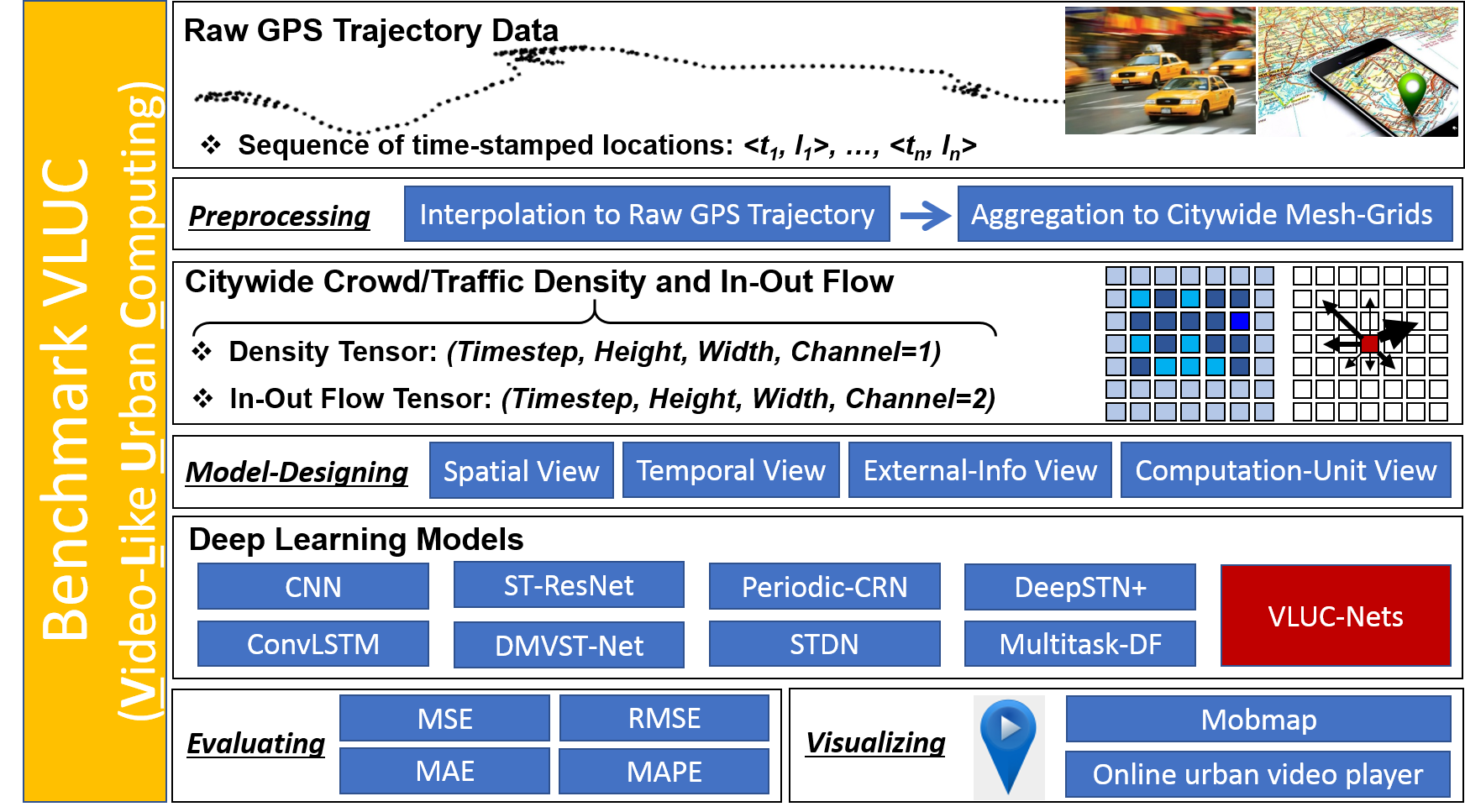}
	\caption{An overview of benchmark \textbf{VLUC} (\underline{\textbf{V}}ideo-\underline{\textbf{L}}ike \underline{\textbf{U}}rban \underline{\textbf{C}}omputing).}
	\label{fig:overview}
\end{figure}

``Urban video'' containing citywide crowd and traffic information are high-dimensional sequential data with high complexity, which naturally drive researchers to design advanced deep learning models to achieve superior performance to classical methodologies.  Although the models target different tasks as shown in Table \ref{tab:modelsummary}, they can be easily modified to apply to one another. Currently, as shown in Table \ref{tab:modeldetail}, the evaluations on this family of methods are still insufficient on the following aspects: (1) some fail to compare with other state-of-the-art models; (2) some are validated only on traffic flow data from taxi or bicycle, not on crowd flow data; (3) some utilize extra data sources such as weather data and POI data; (4) some utilize self-designed objective function; (5) some only utilize RMSE as the evaluation metrics; (6) case studies on some specific regions and times like hot station or residential area in morning rush hour are missing. Thus, in this study, we aim to build a standard benchmark to comprehensively evaluate the state-of-the-art methodologies. Specifically, (1) two classic problems are set as targets: crowd/traffic density prediction and in-out flow prediction \cite{hoang2016forecasting,zhang2016dnn}. The former is to predict how many people/vehicles will be in each mesh-grid at the next timestamp, and the latter is to predict how many people/vehicles will flow into or out from each mesh-grid in next time interval. The task is to take multiple steps of historical observations as input and report the next-step prediction result as output; (2) A new dataset is created using the GPS log data from a popular smartphone app, which can reflect the real-world crowd density and flow. Then the prediction for crowd and traffic can be conducted by using our new dataset and the existing datasets respectively; (3) Unified objective function MSE is adopted for model training, and extra data source and the related processing module are excluded from the models. So that we can fairly verify the pure ability of video-like modeling on spatiotemporal data; (4) Time-series prediction results on selected regions are added as case studies to demonstrate the effectiveness at different places and times. In summary, our work has three-fold contributions as follows:

\begin{itemize}
	\item We propose a new concept called video-like urban computing, and give a comprehensive review of the state-of-the-art works of literature.
	\item We publish a new dataset for crowd density and in-out flow prediction, which is generated based on a real-world smartphone app.
	\item To the best of our knowledge, this is the first attempt to build a standard benchmark that implements multiple state-of-the-art methods, which are thoroughly evaluated on a set of open datasets.
	\item We develop a family of effective models called VLUC-Nets by empirically integrating the advanced deep learning techniques, and validate its performance and robustness on the benchmark.	
\end{itemize}
An overview of the proposed benchmark VLUC has been shown in Fig.\ref{fig:overview}. With this benchmark, we hope researchers can easily follow up and quickly launch a new solution on this topic.

The remainder of this paper is organized as follows. In Section 2, we introduce data preprocessing and give problem definition. In Section 3, we provide a description of the datasets. In Section 4, we explain the implemented models. In Section 5, we present the evaluation results. In Section 6, we briefly recap some other related works. In Section 7, we give our conclusion and discuss future work.

\section{Preliminary}
In this section, we first introduce how to do the data preprocessing and then give the problem definition of density and in-out flow prediction. Without loss of generality, in this paper, we use object to refer to people, vehicle, and bicycle in different GPS data sources.

\begin{figure}[h]
	\centering	
	\includegraphics[width=1.0\textwidth]{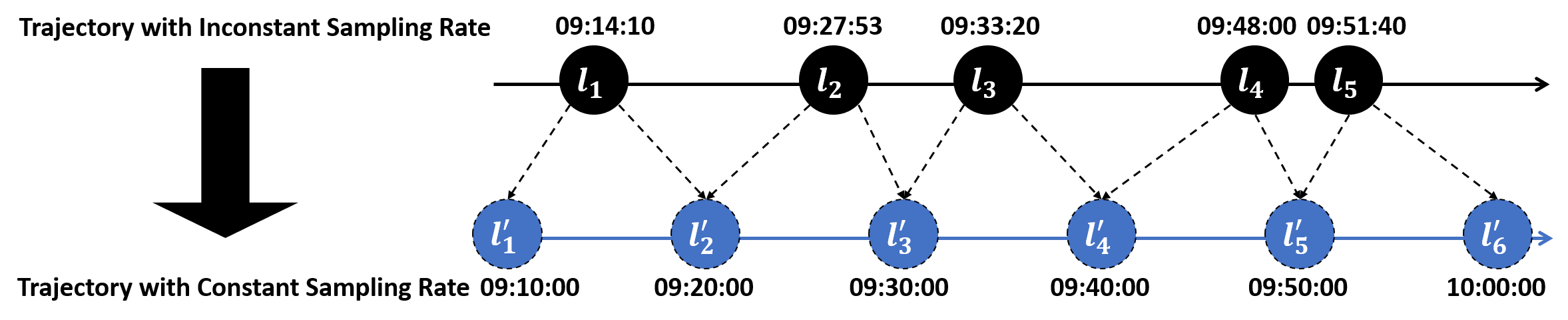}
	\caption{The illustration of data preprocessing is listed, through which the raw trajectory data with inconstant sampling rate will be calibrated to the trajectories with constant sampling rate. For example, when $\Delta{\tau}$ is set to 10 minutes, the raw trajectory snippet \{$l_1$, $l_2$, ..., $l_5$\} timestamped at \{09:14:10, 09:27:53, ..., 09:51:40\} will be converted to a calibrated trajectory snippet \{$l'_1$, $l'_2$, ..., $l'_6$\} constantly timestamped at \{09:10:00, 09:20:00, ..., 10:00:00\}. The coordinate value of $l'_3$ is calculated through linear interpolation based on $l_2$ and $l_3$.}
	\label{fig:preprocessing}
\end{figure}

\subsection{Data Preprocessing}
Object trajectory is represented by a sequence of 3-tuple: ($timestamp$, $latitude$, $longitude$), which be further simplified as a sequence of ($t$, $l$)-pair. Object trajectory is stored and indexed by day ($i$) and object id ($o$) in the database $\Gamma$. Given a time interval $\Delta{\tau}$, each object's trajectory on each day $\Gamma_{io}$ 
is calibrated to obtain constant sampling rate as follows:
\begin{equation*}	
	\Gamma_{io}=(t_1, l_1), ..., (t_k,l_k), \forall j \in \left[1,k\right), |t_{j+1} - t_j| = \Delta{\tau}
\end{equation*}
which has been illustrated in Fig.\ref{fig:preprocessing}.
Then given the mesh-grids of an urban area \{$g_1$,$g_2$,...,$g_{Height*Width}$\}, trajectory $\Gamma_{io}$ is mapped onto mesh-grids as follows:  
\begin{equation*}	
	\Gamma_{io}=(t_1, g_1), ..., (t_k,g_k), \forall j \in \left[1,k\right], l_j \in g_j
\end{equation*}
Lastly, density video and in-out flow video can be aggregated and generated with the processed trajectories with constant sampling rate according to the definitions below.

\subsection{Problem Definition}
\emph{Definition 1} {(Density and In-Out Flow):} Crowd/traffic density at timestamp $t$ in mesh-grid $g_m$ is defined as follows:
\begin{equation*}
	d_{tm} = |\{o | \Gamma_{o}.g_t = g_m \}|
\end{equation*}
According to \cite{hoang2016forecasting,zhang2016dnn}, crowd/traffic in-out flow between consecutive timestamps $t$-1 and $t$ in mesh-grid $g_m$ is defined as follows: 
\begin{equation*}
	f_{tm}^{(in)} = |\{o | \Gamma_{o}.g_{t-1} \ne g_m \wedge \Gamma_{o}.g_{t} = g_m\}|
\end{equation*}	
\begin{equation*}
	f_{tm}^{(out)} = |\{o | \Gamma_{o}.g_{t-1} = g_m \wedge \Gamma_{o}.g_{t} \ne g_m\}|
\end{equation*}	
Then, by representing the mesh-grids with a 2D index ($H$,$W$), density and in-out flow video containing $T$ consecutive frames can be represented by a 4D tensor $\mathbb{R}^{T\times H\times W\times C}$, where channel $C$ for density and in-out flow are equal to 1 and 2 respectively. Here, Min-Max normalization is used to scale the scalar values into [0, 1].

\noindent\emph{Definition 2} {(Density and In-Out Flow Prediction):} Given historical observations of density and in-out flow $x_d$=$d_1, ..., d_{t}$, $x_f$=$f_1, ..., f_{t}$ at timestamp $t$, building prediction models for the next-step density and in-out flow $y_d$=$d_{t+1}$, $y_f$=$f_{t+1}$ is to obtain such parameters $\theta_{d}$, $\theta_{f}$ that can minimize the objective function $\mathcal{L}(\cdot)$ respectively as follows:
\begin{equation*}
	\theta_{d} = \mathop{\argmin}_{\theta_{d}}{\mathcal{L}(\hat{Y}_d, Y_d)}
\end{equation*}
\begin{equation*}
    \theta_{f} = \mathop{\argmin}_{\theta_{f}}{\mathcal{L}(\hat{Y}_f, Y_f)}
\end{equation*}
where Mean Squared Error (MSE) = ||$Y$-$\hat{Y}$||$_2^2$ is adopted as $\mathcal{L}(\cdot)$ in our benchmark.

\section{Dataset}
In this section, we introduce a new dataset and the existing ones used for our benchmark.

\begin{table*}[h]
	\centering
	\caption{DataSet Summary}
	\label{tab:datasummary}
	\begin{tabular*}{15.8cm}{@{\extracolsep{\fill}}|l|l|l|l|l|l|}
		\hline
		Dataset & Data Type & Mesh Size & $H$, $W$ & Time Period, Interval & Max Value\\
		\hline
		BousaiTYO & Density, In-Out Flow& 450m$\times$450m & 80, 80 & 2017/4/1/-2017/7/9, 0.5 hour & 2965, 887\\
		BousaiOSA & Density, In-Out Flow & 450m$\times$450m & 60, 60 & 2017/4/1/-2017/7/9, 0.5 hour & 1998, 435\\
		TaxiBJ & In-Out Flow & unknown & 32, 32& inconsecutive 4 parts, 0.5 hour & 1292 \\
		BikeNYC & In-Out Flow  & unknown & 16, 8 & 2014/4/1-2014/9/30, 1 hour & 267 \\
		BikeNYC-I & In-Out Flow & unknown & 21, 12 & 2014/4/1-2014/9/30, 1 hour & 737 \\
		BikeNYC-II & In-Out Flow  & 1km$\times$1km & 10, 20 & 2016/7/1-2016/8/29, 0.5 hour & 307\\
		TaxiNYC & In-Out Flow & 1km$\times$1km & 10, 20 & 2015/1/1-2015/3/1, 0.5 hour & 1289\\
		\hline
	\end{tabular*}
\end{table*}%

\begin{figure}[h]
	\centering	
	\includegraphics[width=1.0\textwidth]{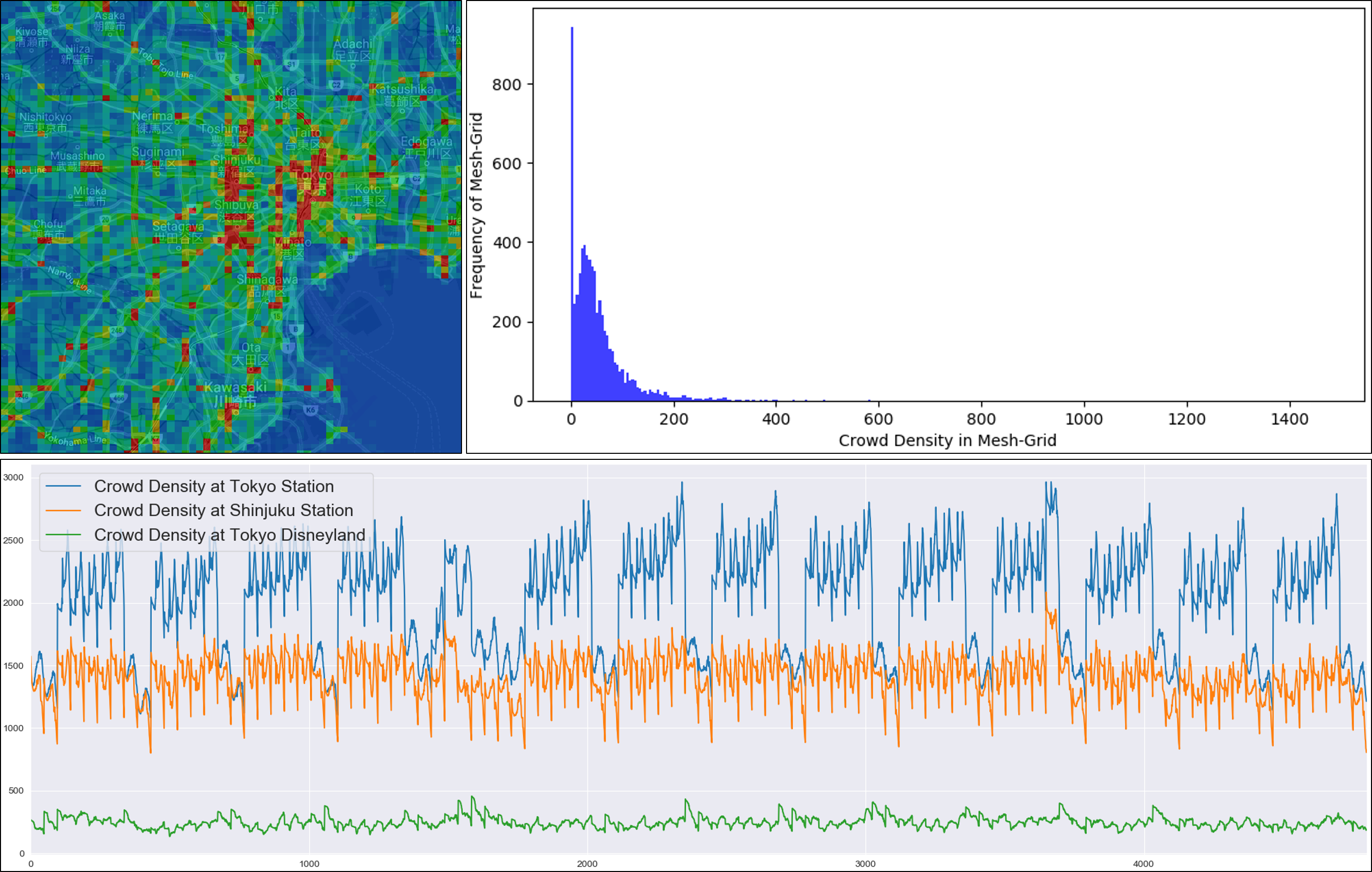}
	\caption{One snapshot of BousaiTYO density at 2017/4/1 09:00:00 is listed on the upper left, where the red color represents the highest density and the blue color represents the lowest density. The crowd density distribution over the 6,400 mesh-grids is plotted on the upper right. Time-series of crowd density over 100 days (4,800 timestamps) in Tokyo station, Shinjuku station, and Tokyo Disneyland is plotted at the bottom.}
	\label{fig:basic}
\end{figure}

\subsection{BousaiTYO and BousaiOSA}
AAAAA (ANONYMIZED IT COMPANY for blind review) provides a smartphone application called Bousaisokuho to give early information and warning towards different disasters such as earthquake, rain, snow, and tsunami. Users are required to provide their location information so that Bousai App can precisely send local disaster alerts to the users in relevant areas. Also, real-time GPS trajectory data are anonymously collected under users' consent for real-time notification and research purposes. GPS logs will be generated when the smartphone user stops staying at one location and starts moving by identifying the change of current location. Every day since 2017, the GPS logs are being generated from around 1 million users (approximately 1\% of the total population of Japan). The file size of each day is about 18 GB, containing approximately 150 million GPS records. Each record includes user ID, timestamp, latitude, and longitude. The sampling rate of each user's GPS data is approximately 20 records per day, which is similar to common call detail records (CDR) data, but sparser than many taxi GPS data. 

We select two big cities in Japan (Tokyo\footnote{Tokyo: $Longitude$ $\in$ [139.50, 139.90], $Latitude$ $\in$ [35.50, 35.82]} and Osaka\footnote{Osaka: $Longitude$ $\in$ [135.35, 135.65], $Latitude$ $\in$ [34.58, 34.82]}) as target urban areas, 100 consecutive days from 2017/4/1/ to 2017/7/9 as target time period. We crop the raw GPS trajectory data in this spatiotemporal range out. As data preprocessing, we first conduct data cleaning and noise reduction to the raw GPS trajectory data, and then do linear interpolation to make sure each user's 24-hour (00:00$\sim$23:59) GPS log has a constant sampling rate with $\Delta{\tau}$ equal to 5 minutes. To generate video-like data, we mesh each area with $\Delta{Lon.}$=0.005 $\Delta{Lat.}$=0.004 (approximately 450m$\times$450m), which results 80$\times$80 and 60$\times$60 mesh-grids respectively for each city. We get one timestamp every 30 minutes, so there are 4800 timestamps (100 days$\times$48/day) in total. According to Definition 1, crowd density and in-out flow are generated for each timestamp. To further eliminate the concerns of user privacy problem, k-anonymization is conducted by setting the scalar values less than 10 to 0. Finally, for Tokyo dataset BousaiTYO, the density and in-out flow video can be represented by tensor (4800, 80, 80, 1) and (4800, 80, 80, 2) respectively;  for Osaka dataset BousaiOSA, the density and in-out flow video can be represented by tensor (4800, 60, 60, 1) and (4800, 60, 60, 2) respectively. A series of visualization results about BousaiTYO density data have been listed as Fig.\ref{fig:basic}.

\begin{figure}[t]
	\centering	
	\includegraphics[width=0.9\textwidth]{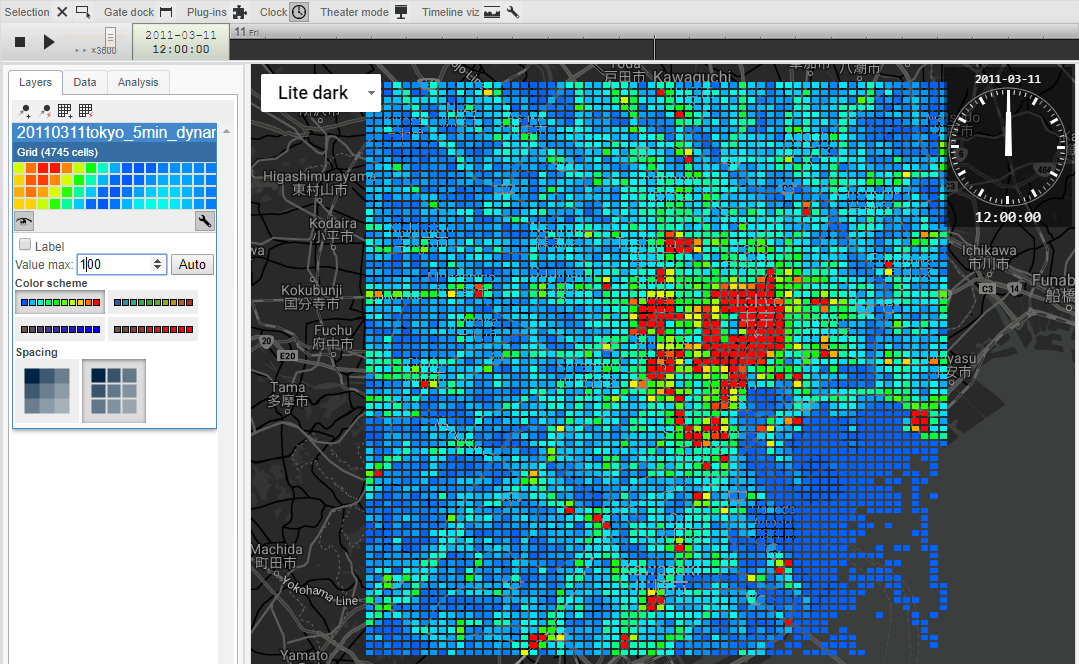}
	\caption{The online visualization tool (Mobmap) runs like a typical video player for urban video data.}
	\label{fig:mobmap}
\end{figure}

Moreover, in our benchmark, we integrate an online visualization tool called Mobmap (ANONYMIZED URL) to demonstrate the datasets, which can function like a typical video player to play urban video data as shown in Fig.\ref{fig:mobmap}. By employing the dynamic grid data visualization module of Mobmap, it displays the crowd density variation of Tokyo frame by frame on the map, just like playing a video. The module receives the geographic grid data (timestamp, meshcode, value) as input, outputs the frame-by-frame image data by timestamp aggregation and color mapping. It enables users to observe the trend of grid data over time as if watching a video, which provides a new possibility to find valuable insights. To improve the flexibility of visualization analysis and enhance the observation experience,  Mobmap further integrates a series of interactive operations. (1) It provides play, pause, and fast-forward buttons, enabling users to observe on various time scales and specific moments, e.g., Fig.4 shows the crowd density statue at 12:00:00 AM. (2) It provides a set of practical color schemes and allows users to adjust the upper bound of the color mapping based on their requirements (i.e., if the grid value is higher than the bound, it will be mapped to a fixed color). e.g., In Fig.\ref{fig:mobmap}, we can efficiently define and monitor the hottest regions over time, we chose the No.1 color scheme with the bound set to 100, which means if the grid crowd density is higher than 100, the mapped color will always be red(hottest region). Users can customize color schema and threshold based on data properties, task requirements, and personal preference, to achieve the most reasonable visualization effect.

\subsection{TaxiBJ, BikeNYC, and TaxiNYC}
\noindent\textbf{TaxiBJ.} This is taxi in-out flow data used by \cite{zhang2017deep,zonoozi2018periodic}, created from the taxicab GPS data in Beijing from four separate time periods: 2013/7/1-2013/10/30, 2014/3/1-2014/6/30, 2015/3/1-2015/6/30, and 2015/11/1-2016/4/10.

\noindent\textbf{BikeNYC.} This is bike in-out flow data used by \cite{zhang2017deep}, taken from the NYC Bike system from 2014/4/1 to 2014/9/30. Similar datasets BikeNYC-I, BikeNYC-II were used by \cite{lin2019deepstn+} and \cite{yao2019revisiting} respectively. These two will be integrated into our benchmark since they have a larger spatial range than BikeNYC.

\noindent\textbf{TaxiNYC.} This is taxi in-out flow data used by \cite{yao2019revisiting}, created from the NYC-Taxi data in 2015.

The details of our new dataset, as well as the existing ones, are given in Table \ref{tab:datasummary}. Through it, we can see the advantage of our new dataset on the following aspects: (1) large urban area; (2) fine-grained mesh size; (3) high user sampling rate. Our benchmark will integrate BousaiTYO-OSA for crowd density and in-our flow prediction task; TaxiBJ, BikeNYC I-II, and TaxiNYC for traffic in-out flow prediction task.

\section{Model}
In this section, we introduce the models implemented in our benchmark for comparison and evaluation.

\subsection{Baselines and The-State-of-The-Arts}
\noindent\textbf{HistoricalAverage.} Density and in-out flow for each timestamp are estimated by averaging the historical values from the corresponding timestamp in the training dataset, and weekday and weekend will be considered separately.

\noindent\textbf{CopyYesterday.} We directly copy the corresponding observation (frame) from the previous day (yesterday) as the result. 




\noindent\textbf{CNN.} It is a basic deep learning predictor constructed with four CNN layers. The 4D tensor would be converted to 3D tensor ($H$,$W$,$T$*$C$) by concatenating the channels at each timestep just like the way \cite{zhang2017deep} did, so that CNN could take a 4D tensor as input. The CNN predictor utilizes four Conv layers to take the current observed $t$-step frames as input and predicts the next frame as output. The four Conv layers use a 32 filters of 3$\times$3 kernel window, and the fourth Conv layer uses a ReLU activation function to output the next frame (step) of urban video. BatchNormalization is added between two consecutive layers.

\noindent\textbf{ConvLSTM.} Convolutional LSTM \cite{xingjian2015convolutional} extends the fully connected LSTM (FC-LSTM) to have convolutional structures in both the input-to-state and state-to-state transitions. ConvLSTM has achieved new success on video modeling tasks due to its superior performance in capturing both spatial and temporal dependency than CNN. Thus, a stronger predictor can be constructed with four ConvLSTM layers, which also takes current $t$-step observations as input and predicts the next step. The ConvLSTM layers use a 32 filters of 3$\times$3 kernel window and the ReLU activation is used in the final layer. BatchNormalization is also added between two consecutive layers.

\noindent\textbf{Multitask-DF}. Multitask learning \cite{ngiam2011multimodal} was employed by \cite{zhang2019flow} and \cite{jiang2019deepurbanevent} to model two correlated tasks together and gain concurrent enhancement. Similarly, we design a ConvLSTM-based multitask model to jointly model density and in-out flow video. The motivation comes from the following two points: (1) People tend to follow the trend. Crowded places may attract more and more people to visit; (2) Higher inflow will lead to higher density, higher outflow will lead to lower density. Multitask-DF first takes $X_d$ ($t$-step observed density) and $X_f$ ($t$-step observed in-out flow) with two separate ConvLSTM layers; then concatenates two separate latent representations and passes it to two consecutive ConvLSTM layers; finally outputs $\hat{Y}_d$ (next-step density) and $\hat{Y}_f$ (next-step in-out flow) with two separate ConvLSTM layers. The model parameters $\theta$ can be trained by minimizing the objective function $\mathcal{L}(\cdot)$ as follows: 
\begin{equation*}\label{encodererror}
	\theta = \mathop{\argmin}_{\theta}{[\lambda\mathcal{L}(\hat{Y}_d, Y_d) + (1-\lambda)\mathcal{L}(\hat{Y}_f, Y_f)]}
\end{equation*}
where $\lambda$ is fine-tuned as 0.3 in the experiment. \textbf{ConvLSTM} and \textbf{Multitask-DF} naturally rely on the superior performance of ConvLSTM to capture both spatial and temporal dependency. 

\begin{figure}[t]
	\centering
	\includegraphics[width=1.0\textwidth]{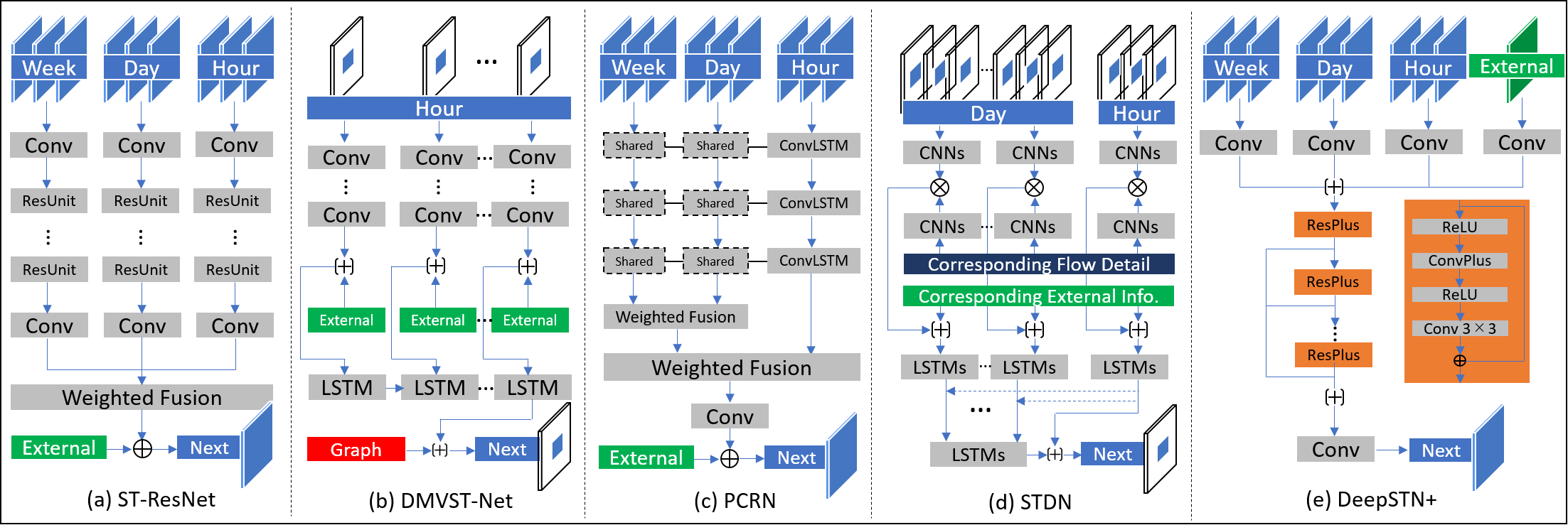}
	\caption{Model Architecture Summary.}
	\label{fig:architectures}
\end{figure}
Keeping the same setting with \textbf{CNN} and \textbf{ConvLSTM}, every ConvLSTM layer uses a 32 filters of 3$\times$3 kernel window, and the ReLU activation is added only in the final layer.
 
The core architectures of other state-of-the-art models from ST-ResNet to DeepSTN+ are simplified and summarized as Fig.\ref{fig:architectures}. They will be comprehensively and uniformly reviewed from the following aspects: (1) core technique to capture spatial and temporal dependency; (2) input feature; (3) computation unit; (4) extra component.

\noindent\textbf{ST-ResNet}. ST-ResNet\cite{zhang2017deep} is a CNN-based deep learning method for traffic in-out flow prediction. To capture citywide spatial dependency, it employs residual learning to construct deep enough CNN networks; To capture temporal dependency, it first designs a set of unique features namely $Closeness$,$Period$, and $Trend$, which correspond to \emph{the recent time intervals}, \emph{daily periodicity}, and \emph{weekly trend} respectively, then fuses them together through three learnable parametric matrices. Intuitively, the three sequences can be represented by [$X_{t-l_c}$, $X_{t-(l_c-1)}$, ..., $X_{t-1}$], [$X_{t-l_p \cdot p}$, $X_{t-(l_p-1) \cdot p}$, ..., $X_{t-p}$], and [$X_{t-l_q \cdot q}$, $X_{t-(l_q-1) \cdot q}$, ..., $X_{t-q}$], where $l_c$, $l_p$, $l_q$ are the sequence length of $Closeness$,$Period$,$Trend$, $p$ and $q$ are the span of $Period$ and $Trend$, the $Closeness$ span is equal to 1 by default. 4D tensor would be converted to 3D tensor ($H$,$W$,$T$*$C$) by concatenating the channels at each timestep. The computation unit is the whole citywide image. Additionally, it further utilizes weather, holiday event information, and metadata(i.e. DayOfWeek, Weekday/Weekend) as external information. To verify the pure ability of capturing spatial and temporal dependency, only metadata will be utilized in our benchmark. 

\noindent\textbf{DMVST-Net}. DMVST-Net\cite{yao2018deep} is a deep learning method for taxi demand prediction based on CNN and LSTM. It uses a local CNN to capture spatial dependency only among nearby grids; employs LSTM to capture temporal dependency only from the recent time intervals (i.e. $Closeness$). The local CNN takes one grid and its surrounding grids (i.e. $S$$\times$$S$ region) as the input, and a separate and unshared CNN is constructed for each timestamp. The input tensor is essentially ($T$,$S$,$S$,$1$), and the computation unit is grid (pixel) rather than citywide image. Furthermore, it constructs a weighted graph, where nodes are the grids, and each edge represents the similarity of two time-series values (i.e. historical taxi demand) between any two grids; then it embeds this graph into a feature vector and concatenates it with the main feature vector from LSTM layer. Through this, it can improve the ability to capture citywide spatial dependency. Similarly, in our benchmark, only metadata will be utilized as external information.

\noindent\textbf{PCRN}. Periodic-CRN\cite{zonoozi2018periodic} is a ConvGRU-based deep learning model for taxi density and in-out flow prediction by fully making use of recurrent periodic patterns. To capture citywide spatial dependency, it builds a pyramidal architecture by stacking three convolutional RNN layers. To capture temporal dependency, it first learns a representation from the observations of $Closeness$ through the stacked pyramidal ConvGRUs; it divides the representations into two types of periodic patterns, namely daily and weekly pattern, each of them is a set of periodic representations corresponding to a specific time span (i.e. day or week); then it maintains a memory-based dictionary to reuse and update the two types of periodic patterns dynamically; lastly it employs a weighting based fusion to merge periodic representations with the current representation of input sequence. Thus, the input feature can be seen as $Closeness$, $Period$, and $Trend$. The computation unit is the whole citywide image. Also, only metadata will be utilized as external information. In our benchmark, we replace ConvGRU with ConvLSTM and simplify the architecture as Fig.\ref{fig:architectures}-(c) shows.


\noindent\textbf{STDN.} Spatial-Temporal Dynamic Network (STDN)\cite{yao2019revisiting} is an improved version of DMVST-Net for taxi/bike Origin-Destination number (volume) prediction. To capture spatial dependency, it inherits the local CNN technique from DMVST-Net, and further designs a flow gating mechanism to fuse local flow information (i.e. flow from one central grid to its surrounding $S$$\times$$S$ grids) with the traffic volume information together. In terms of temporal dependency, it improves DMVST-Net by taking not only $Closeness$ information but also long-term daily periodicity (i.e. $Period$) into account. Moreover, it considers the temporal shifting problem about periodicity (i.e. traffic data is not strictly periodic) and designs a \emph{Periodically Shifted Attention Mechanism} to solve the issue. Specifically, it sets a small time window to collect $Q$ time intervals right before and after the currently-predicting one. And it uses an LSTM plus attention mechanism to obtain a weighted average representation $h$ from the time intervals in each window. For previous $P$ days to be considered as $Period$, it gets a sequence of representations ($h_1$, $h_2$, $...$, $h_P$), then it uses another LSTM layer to extract the final periodic representation from the sequence. The computation unit is grid (pixel) same with DMVST-Net. Lastly it jointly models inflow (start traffic volume) and outflow (end traffic volume) together. The flow gating mechanism will be pruned in our benchmark since flow detail information are needed.

\noindent\textbf{DeepSTN+.} DeepSTN+ \cite{lin2019deepstn+} is an improved version of ST-ResNet for crowd and traffic in-out flow prediction. It directly inherits the input features (i.e. $Closeness$, $Period$, and $Trend$), and enhances the ST-ResNet from the following aspects: (1) to capture longer-range spatial dependency, it designs a unique $ConvPlus$ block, and replaces the ordinary $Conv$-based residual unit in ST-ResNet with $ResPlus$ (i.e. $ConvPlus$-based residual unit). Furthermore, multi-scale fusion mechanism is employed to preserve the representation from each $ResPlus$ layer; (2) in terms of temporal dependency, it applies early-fusion mechanism instead of end-fusion in ST-ResNet to get better interaction among $Closeness$,$Period$, and $Trend$; (3) additionally, it takes the influence of location function on the crowd/traffic flow into consideration by using POI data to gain a semantic plus. The computation unit is the whole citywide image. Similarly, we prune the POI processing component from DeepSTN+ to verity the pure spatiotemporal modeling capability.


\begin{figure}[t]
	\centering
	\includegraphics[width=1.0\textwidth]{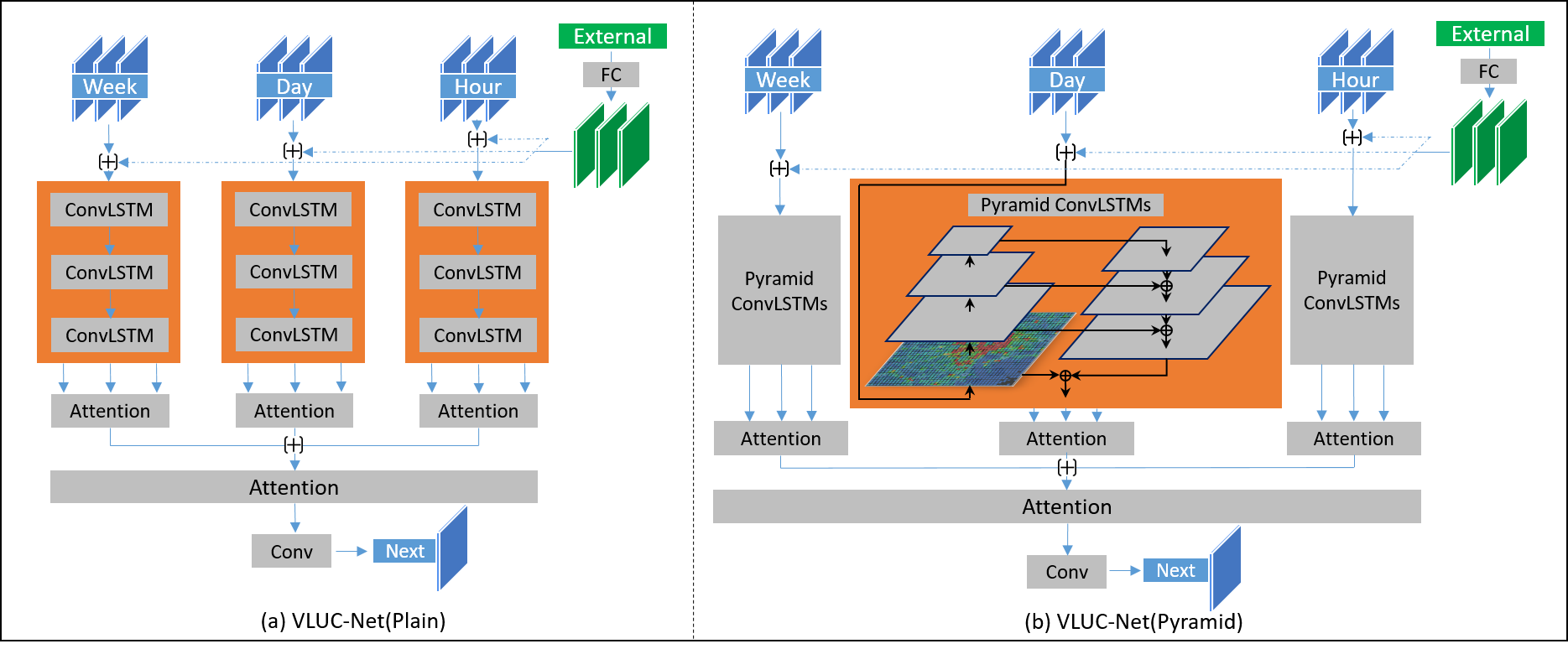}
	\caption{Architecture of VLUC-Nets: VLUC-Nets with plain ConvLSTMs and VLUC-Nets with pyramid ConvLSTMs.}
	\label{fig:vlucnets}
\end{figure}

\subsection{VLUC-Nets}
On this benchmark, we propose a novel family of deep learning models named after our benchmark: VLUC-Nets. Through the experience of implementing the state-of-the-arts on the benchmark, we notice that the models with pixel (mesh-grid) as the computing unit will not be so efficient, especially for a large spatial domain like BousaiTYO ($H$,$W$=$80$,$80$) or BousaiOSA ($H$,$W$=$60$,$60$). Therefore, we first set 
the computation unit as the whole citywide image for VLUC-Nets. Moreover, ConvLSTM has demonstrated the advantage on simultaneously capturing the spatial and temporal dependency, thus we build VLUC-Nets by utilizing ConvLSTM as the basic component like the way \textbf{PCRN} and \textbf{Multitask-DF} did. We inherit and modify the wide used input features $Closeness$, $Period$, and $Trend$ as [$X_{(t-T_c)-l_c}$, $X_{(t-T_c)-(l_c-1)}$, ..., $X_{(t-T_c)-1}$], [$X_{(t-T_p)-l_c}$, $X_{(t-T_p)-(l_c-1)}$, ..., $X_{(t-T_p)-1}$], and [$X_{(t-T_t)-l_c}$, $X_{(t-T_t)-(l_c-1)}$, ..., $X_{(t-T_t)-1}$], where $T_c$, $T_p$, and $T_t$ represent the periodicity of $Closeness$, $Period$, and $Trend$. 
On Bousai dataset where the time interval is 30 minutes, when $T_c$ is set to 0, $Closeness$ corresponds to the current momentary observations; when $T_p$ is set to 48, $Period$ corresponds to the observations from previous day (i.e., daily periodicity); when $T_t$ is set to $7*48$, $Trend$ corresponds to the observations from previous week (i.e., weekly periodicity). Also, only date information will be utilized as external (meta) information. Finally, after consolidating the basic strategies, VLUC-Nets are elaborately built based on ConvLSTM following the standard model-designing views as shown in Fig.\ref{fig:overview}. Two versions of VLUC-Nets are implemented in our benchmark as shown in Fig.\ref{fig:vlucnets}. The plain version is designed for datasets with small spatial domain, while the pyramid version is designed for datasets with large spatial domain. The concrete techniques are listed as follows:

\begin{itemize}
	\item \textbf{Temporal View.} Attention is all we need\cite{vaswani2017attention}, and LSTM plus attention mechanism has been seen as a state-of-the-art technique in sequential modeling tasks. Especially, in the urban computing filed, LSTM plus attention has achieved a lot of success on individual's next-location prediction (DeepMove\cite{feng2018deepmove}), traffic time prediction for each road path (DeepTTE\cite{wang2018will}), and traffic volume and flow prediction (STDN\cite{yao2019revisiting}). Here, we employ and extend the attention mechanism from LSTMs to ConvLSTMs. Originally, for LSTMs, an attention block takes a 3D tensor ($Batch$, $Timestep$, $Feature$) as input, and outputs a 2D attention tensor ($Batch$, $Feature$). Now, for ConvLSTMs, the attention block takes a 5D tensor ($Batch$, $Timestep$, $Height$, $Width$, $Channel$) as input, and outputs a 4D attention tensor ($Batch$, $Height$, $Width$, $Channel$). 
	The formulas for the attention block are listed as follows:
	\begin{equation*}
	h_{att} = \sum_{i=1}^{l_c}\alpha_i \cdot h_{i}
	\end{equation*}
	\begin{equation*}
	z_i = tanh(W*h_{i} + b)
	\end{equation*}
	\begin{equation*}
	\alpha_i = \frac{e^{z_i}}{\sum_{j}{e^{z_j}}}
	\end{equation*}
	Here, $W$ is the weights of fully-connected (FC) layers for $i$-th hidden state in $H$ outputted by ConvLSTM layers. The hidden states of ConvLSTMs $H=\{h_{1},h_{2},...,h_{l_c}\}$ will be fused as one attention state denoted as $h_att$.
	For each set of ConvLSTMs $Closeness$, $Period$, and $Trend$, each individual attention block will generate an attention state, namely $h_{att_c}$, $h_{att_p}$, and $h_{att_t}$. Then the list of attention states \{$h_{att_c}$, $h_{att_p}$, $h_{att_t}$\} will be further fed into another attention block to fuse the $Closeness$, $Period$, and $Trend$ information together as shown in Fig.\ref{fig:vlucnets}-(a).
	\item \textbf{Spatial View.} Extracting and utilizing pyramid and hierarchical features are taken as the state-of-the-art technique in the computer vision-related literatures such as Feature Pyramid Networks (FPN) \cite{Lin_2017_CVPR}, Mask R-CNN \cite{He_2017_ICCV}, and Faster R-CNN \cite{NIPS2015_5638}. To better capture the citywide spatial dependency, pyramid ConvLSTMs are utilized to replace the plain-stacked ConvLSTMs as shown in Fig.\ref{fig:vlucnets}-(b), where the outputs from the lower ConvLSTM layers are concatenated with the upper layers. The latent representations derived from pyramid ConvLSTM are considered to contain richer spatial information than the ones from plain ConvLSTMs. In addition, pyramid ConvLSTMs can shorten the distance from the input to output which functions similarly as deep residual networks, but pyramid networks do not require to build a very deep network to gather the upper-level features from the input.
	\item \textbf{External-Info View.} In spatio-temporal city computing applications, external information such as timestamps and other metadata also have significant influences on model performance. Instead of fusing them with fully connected layer or simple convolutional layer, we employed ConvLSTM to explore the complex interactions between spatio-temporal data and metadata. Specifically, external information, including time and day, day of week and holiday flag, are sent to one fully connected layer after one-hot encoding, the output will be reshaped to a 5D tensor ($Batch$, $Timestep$, $Height$, $Width$, $Channel$) and concatenated with spatio-temporal data to obtain the input of Pyramid ConvLSTMs. This early input mechanism for external enables the model to deeper fuse multiple heterogeneous data and faster convergence.
\end{itemize}
	
	

\section{Experiment}
In this section, we present the evaluation results for the state-of-the-art models as well as baseline models.

\begin{figure}[h]
	\centering
	\includegraphics[width=0.9\textwidth]{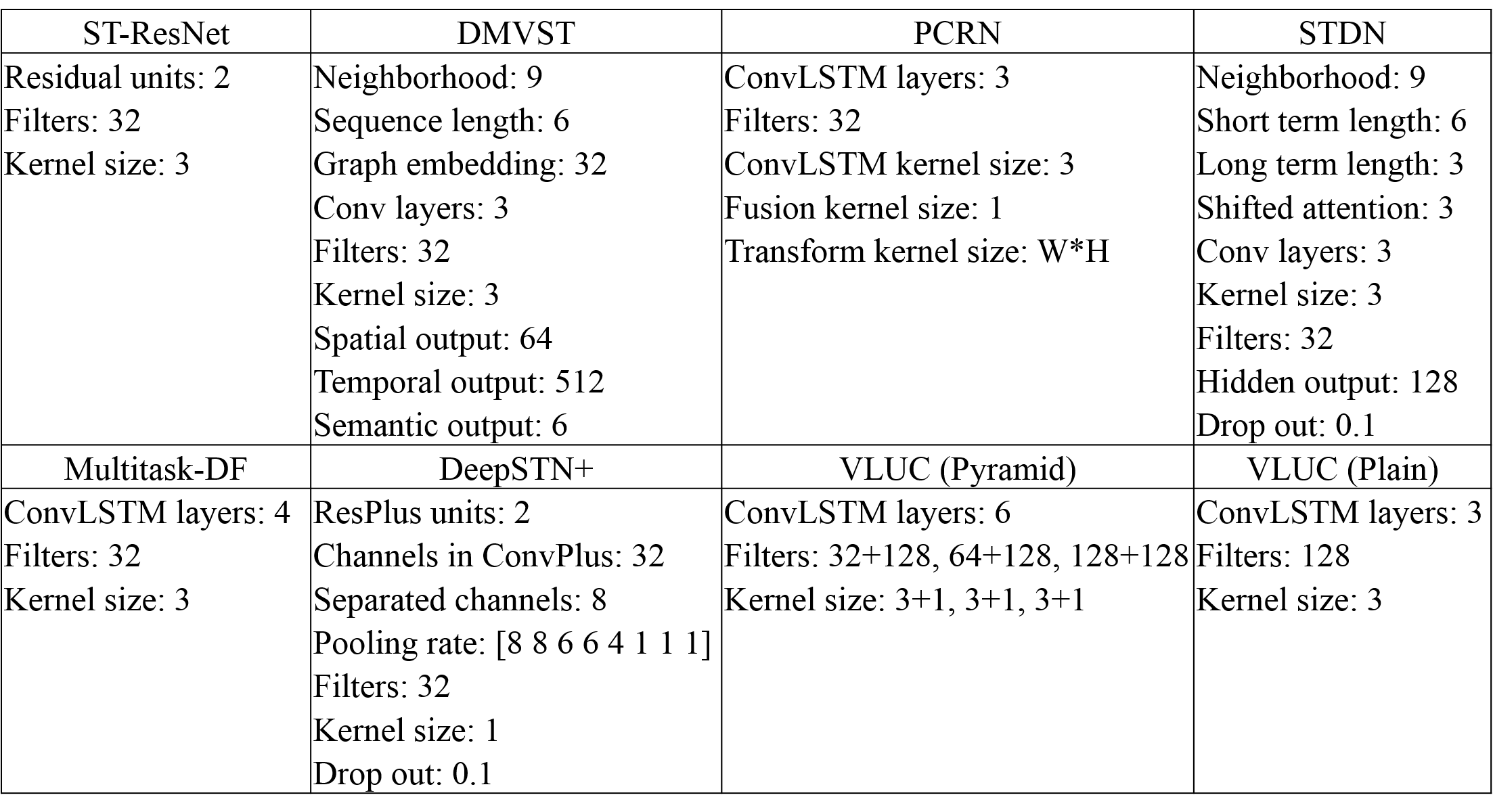}
	\caption{Summary of fine-tuned hyper-parameters. Note that the parameters displayed in the list [8,8,6,6,4,1,1,1] correspond to the settings for the dataset list [BousaiTYO Density, ..., BikeNYC-I, BikeNYC-II].}
	\label{fig:hyperparameter}
\end{figure}

\subsection{Parameter Setting} 
The following settings were kept the same for each model and dataset. We set the current observation step to 6 (i.e. $Closeness$), and used the corresponding observations from previous 1 day and 1 week as $Period$ and $Trend$ respectively, which means $l_c$, $T_c$, $T_p$, and $T_t$ are set to \{6, 0, 7, 7*48\} respectively. It should be noted that these settings on $Closeness$,$Period$,$Trend$ follow the ones widely used and well tuned in the state-of-the-arts listed in Table \ref{tab:modelsummary}. Data from the first 80\% were set as training data (20\% of which were taken as validation data), and the other 20\% were set as testing data. Adam was employed to control the overall training process, where the batch size was set to 4 and the learning rate was set to 0.0001. The training algorithm would either be early-stopped if the validation error converged within 10 epochs or be stopped after 200 epochs, and the best model on validation data would be saved. We rescaled the predicted value back to the original value. All models were run multiple times on each dataset, and the best result would be recorded. Our benchmark was built based upon Keras\cite{keras} and TensorFlow\cite{tensorflow}. The experiments were performed on 3 GPU servers, together with six GeForce GTX 1080Ti graphics cards and two Tesla P40 graphics cards. For all approaches, we used gird search to tune the parameters based on the validation dataset and chose the best set of hyperparameters. We also considered the parameter settings recommended by the original study. The hyper-parameters are finely tuned as Fig.\ref{fig:hyperparameter} by spending equal amounts of effort in each.

\subsection{Evaluation Metric} 
We evaluate the effectivenesses of the models with MSE (Mean Squared Error), RMSE (Root Mean Square Error), MAE (Mean Absolute Error), and MAPE (Mean Absolute Percentage Error):
\begin{equation*}\label{MSE} 
    MSE = \frac{1}{n} \sum_{i}^n ||\hat{Y}_i-Y_i||^{2}
\end{equation*}
\begin{equation*}\label{RMSE} 
	RMSE = \sqrt{\frac{1}{n} \sum_{i}^n ||\hat{Y}_i-Y_i||^{2}}
\end{equation*}
\begin{equation*}\label{MAE} 
	MAE = \frac{1}{n} \sum_{i}^n |\hat{Y}_i-Y_i|
\end{equation*}
\begin{equation*}\label{MAPE} 
MAPE = \frac{1}{n} \sum_{i}^n |\frac{\hat{Y}_i-Y_i}{Y_i}|
\end{equation*}
where $n$ is the number of samples, $Y$ and $\hat{Y}$ are the ground-truth tensor and predicted tensor.
 
\begin{table*}[h]
	\centering
	\caption{Effectiveness Evaluation of Crowd Density and In-Out Flow Prediction on BousaiTYO}
	\label{tab:bousaidata1}
	\begin{tabular*}{15cm}{@{\extracolsep{\fill}}|l|c|c|c|c|c|c|c|c|}
		\hline
		\multicolumn{1}{|l|}{} &
		\multicolumn{4}{c|}{Tokyo Density} &
		\multicolumn{4}{c|}{Tokyo In-Out Flow}
		\\
		\hline
		\multicolumn{1}{|l|}{Model} & 
		\multicolumn{1}{c|}{MSE} & 
		\multicolumn{1}{c|}{RMSE} & 
		\multicolumn{1}{c|}{MAE} &
		\multicolumn{1}{c|}{MAPE} & 
		\multicolumn{1}{c|}{MSE} &
		\multicolumn{1}{c|}{RMSE} & 
		\multicolumn{1}{c|}{MAE} & 
		\multicolumn{1}{c|}{MAPE}
		\\
		\hline
		HistoricalAverage & 190.162 & 13.790 & 6.866 & 3.72\% & 38.183 & 6.179 & 1.855 & 4.15\% \\				
		CopyYesterday & 963.744 & 31.044 & 11.531 & 5.11\% & 75.911 & 8.713 & 2.532 & 4.65\%\\
		CNN & 78.642 & 8.868 & 3.787 & 2.19\% & 42.896 & 6.550 & 1.965 & 4.03\% \\
		ConvLSTM & 61.363 & 7.833 & 3.127 & 1.85\% & 14.314 & 3.783 & 1.440 & 3.66\% \\
		ST-ResNet & 39.900 & 6.317 & 3.459 & 2.05\% & 11.328 & 3.366 & 1.412 & 3.39\% \\
		DMVST-Net & 24.194 & 4.919 & 2.656 & 1.59\% & 21.645 & 4.652 & 1.686 & 3.98\% \\
		PCRN & 31.192 & 5.585 & 3.296 & 2.14\% & 17.602 & 4.195 & 1.439 & 3.51\% \\
		DeepSTN+ & 60.937 & 7.806 & 3.888 & 2.91\% & 10.764 & 3.281 & 1.331 & 3.23\%\\
		STDN & 23.187 & 4.815 & 2.606 & 1.59\% & 10.733 & 3.276 & 1.472 & 3.56\% \\
		Multitask-DF & 38.095 & 6.172 & 2.531 & 1.77\% & 11.668 & 3.416 & 1.406 & 3.66\%\\
		VLUC-Nets (Pyramid) & \textbf{20.536} & \textbf{4.532} & \textbf{2.432} & \textbf{1.49\%} & \textbf{9.760} & \textbf{3.124} & \textbf{1.170} & \textbf{2.97\%}\\
		\hline
	\end{tabular*}
\end{table*}

\begin{table*}[h]
	\centering
	\caption{Effectiveness Evaluation of Crowd Density and In-Out Flow Prediction on BousaiOSA}
	\label{tab:bousaidata2}
	\begin{tabular*}{15cm}{@{\extracolsep{\fill}}|l|c|c|c|c|c|c|c|c|}
		\hline
		\multicolumn{1}{|l|}{} &
		\multicolumn{4}{c|}{Osaka Density } &
		\multicolumn{4}{c|}{Osaka In-Out Flow} 
		\\
		\hline
		\multicolumn{1}{|l|}{Model} & 
		\multicolumn{1}{c|}{MSE} & 
		\multicolumn{1}{c|}{RMSE} & 
		\multicolumn{1}{c|}{MAE} &
		\multicolumn{1}{c|}{MAPE} & 
		\multicolumn{1}{c|}{MSE} &
		\multicolumn{1}{c|}{RMSE} & 
		\multicolumn{1}{c|}{MAE} & 
		\multicolumn{1}{c|}{MAPE} 
		\\
		\hline
		HistoricalAverage & 74.752 & 8.646 & 4.826 & 4.39\% & 11.510 & 3.393 & 0.920 & 2.73\%\\				
		CopyYesterday & 234.461 & 15.312 & 7.375 & 5.80\% & 19.417 & 4.406 & 1.116 & 2.89\%\\
		CNN & 24.017 & 4.901 & 2.396 & 2.20\% & 50.530 & 7.108 & 1.582 & 3.55\%\\
		ConvLSTM & 18.646 & 4.318 & 2.062 & 2.03\% & 6.101 & 2.470 & 0.779 & 2.45\%\\
		ST-ResNet & 16.522 & 4.065 & 2.292 & 2.09\% & 4.976 & 2.231 & 0.721 & 2.21\%\\
		DMVST-Net & 10.571 & 3.251 & 1.850 & 1.77\% & 8.375 & 2.894 & 0.879 & 2.50\%\\
		PCRN & 13.549 & 3.681 & 2.175 & 2.22\% & 6.896 & 2.626 & 0.736 & 2.35\%\\
		DeepSTN+ & 20.762 & 4.557 & 2.436 & 2.80\% & 5.653 & 2.378 & 0.872 & 2.29\%\\
		STDN & 11.950 & 3.457 & 1.891 & 1.86\% & 4.906 & 2.215 & 0.801 & 2.21\%\\
		Multitask-DF & 13.580 & 3.685 & 1.856 & 1.88\% & 5.336 & 2.310 & 0.731 & 2.32\% \\
		VLUC-Nets (Pyramid) & \textbf{10.145} & \textbf{3.185} & \textbf{1.760} & \textbf{1.72\%} & \textbf{4.887} & \textbf{2.211} & \textbf{0.662} & \textbf{2.18\%}\\
		\hline
	\end{tabular*}
\end{table*}

\begin{table*}[h]
	\centering
	\caption{Effectiveness Evaluation of Traffic In-Out Flow Prediction on TaxiBJ and TaxiNYC}
	\label{tab:opendata1}
	\begin{tabular*}{15cm}{@{\extracolsep{\fill}}|l|c|c|c|c|c|c|c|c|}
		\hline
		\multicolumn{1}{|l|}{} &
		\multicolumn{4}{c|}{TaxiBJ} &
		\multicolumn{4}{c|}{TaxiNYC} 
		\\
		\hline
		\multicolumn{1}{|l|}{Model} & 
		\multicolumn{1}{c|}{MSE} & 
		\multicolumn{1}{c|}{RMSE} & 
		\multicolumn{1}{c|}{MAE} &
		\multicolumn{1}{c|}{MAPE} & 
		\multicolumn{1}{c|}{MSE} &
		\multicolumn{1}{c|}{RMSE} & 
		\multicolumn{1}{c|}{MAE} & 
		\multicolumn{1}{c|}{MAPE}
		\\
		\hline
		HistoricalAverage & 2025.328 & 45.004 & 24.475 & 8.04\% & 463.763 & 21.535 & 7.121 & 4.56\%\\				
		CopyYesterday & 1998.375 & 44.703 & 22.454 & 8.14\% & 1286.035 & 35.861 & 10.164 & 5.78\%\\
		CNN & 554.615 & 23.550 & 13.797 & 8.46\% & 280.262 & 16.741 & 6.884 & 8.08\%\\
		ConvLSTM & 370.448 & 19.247 & 10.816 & 5.61\% & 147.447 & 12.143 & 4.811 & 5.16\% \\
		ST-ResNet & 349.754 & 18.702 & 10.493 & 5.19\% & 133.479 & 11.553 & 4.535 & 4.32\%\\
		DMVST-Net & 415.739 & 20.389 & 11.832 & 5.99\% & 185.601 & 13.605 & 4.928  & 4.49\%\\
		PCRN & 355.511 & 18.855 & 10.926 & 6.24\% & 181.091 & 13.457 & 5.453 & 6.27\%\\
		DeepSTN+ & 329.080 & 18.141 & 10.126 & 5.14\% & 130.407 & 11.420 & 4.441 & 4.45\%\\
		STDN & \textbf{317.764} & \textbf{17.826} & \textbf{9.901} & \textbf{4.81\%} & 126.615& 11.252 & 4.474 & \textbf{4.09\%}\\
		VLUC-Nets (Plain) & 337.754 & 18.378 & 10.325 & 5.40\% & \textbf{113.514} & \textbf{10.654} & \textbf{4.157} & 4.54\%\\
		\hline
	\end{tabular*}
\end{table*}

\begin{table*}[h]
	\centering
	\caption{Effectiveness Evaluation of Traffic In-Out Flow Prediction on BikeNYC-I and BikeNYC-II}
	\label{tab:opendata2}
	\begin{tabular*}{15cm}{@{\extracolsep{\fill}}|l|c|c|c|c|c|c|c|c|}
		\hline
		\multicolumn{1}{|l|}{} &
		\multicolumn{4}{c|}{BikeNYC-I} &
		\multicolumn{4}{c|}{BikeNYC-II} 
		\\
		\hline
		\multicolumn{1}{|l|}{Model} & 
		\multicolumn{1}{c|}{MSE} & 
		\multicolumn{1}{c|}{RMSE} & 
		\multicolumn{1}{c|}{MAE} &
		\multicolumn{1}{c|}{MAPE} & 
		\multicolumn{1}{c|}{MSE} &
		\multicolumn{1}{c|}{RMSE} & 
		\multicolumn{1}{c|}{MAE} & 
		\multicolumn{1}{c|}{MAPE}
		\\
		\hline
		HistoricalAverage & 245.743 & 15.676 & 4.882 & 5.45\% & 23.757 & 4.874 & 1.500 & 3.30\%\\				
		CopyYesterday & 241.681 & 15.546 & 4.609 & 5.36\% & 54.054 & 7.352 & 1.995 & 3.94\%\\
		CNN & 145.549 & 12.064 & 4.088 & 5.82\% & 20.351 & 4.511 & 1.574 & 3.98\%\\
		ConvLSTM & 43.765 & 6.616 & 2.412 & 3.90\% & 10.076 & 3.174 & 1.133 & 2.90\%\\
		ST-ResNet & 37.279 & 6.106 & 2.360 & 3.72\% & 10.182 & 3.191 & 1.169 & 2.86\%\\
		DMVST-Net & 63.849 & 7.990 & 2.833  & 3.93\% & 12.397 & 3.521 & 1.287 & 2.97\%\\
		PCRN & 130.878 & 11.440 & 3.790 & 4.90\% & 10.893 & 3.300 & 1.185 & 3.04\%\\
		DeepSTN+ & 38.497 & 6.205 & 2.489 & 3.48\% & 10.275 & 3.205 & 1.245 & 2.80\%\\
		STDN & \textbf{33.445} & \textbf{5.783} & 2.410 & \textbf{3.35\%} & \textbf{9.022} & \textbf{3.004} & 1.167 & \textbf{2.67\%}\\
		VLUC-Nets (Plain) & 33.997 & 5.831 & \textbf{2.175} & 3.51\% & 9.725 & 3.119 & \textbf{1.124} & 2.89\%\\
		\hline
	\end{tabular*}
\end{table*}

\begin{table*}[h]
	\centering
	\caption{Effectiveness Evaluation of Crowd Density Prediction at Tokyo Station (BousaiTYO Density, RMSE)}
	\label{tab:tokyostation}
	\begin{tabular*}{15cm}{@{\extracolsep{\fill}}|l|c|c|c|c|c|c|c|c|}
		\hline
		\multicolumn{1}{|l|}{} &
		\multicolumn{4}{c|}{Weekday} &
		\multicolumn{4}{c|}{Weekend} 
		\\
		\hline
		\multicolumn{1}{|l|}{Model} & 
		\multicolumn{1}{c|}{08:00} & 
		\multicolumn{1}{c|}{12:00} &
		\multicolumn{1}{c|}{16:00} & 
		\multicolumn{1}{c|}{20:00} & 
		\multicolumn{1}{c|}{08:00} & 
		\multicolumn{1}{c|}{12:00} & 
		\multicolumn{1}{c|}{16:00} & 
		\multicolumn{1}{c|}{20:00} 
		\\
		\hline
		HistoricalAverage&37.357&117.929&164.000&84.214&13.133&42.267&7.800&22.200\\
		CopyYesterday&23.000&213.000&234.000&127.000&761.000&844.000&998.000&1068.000\\
		CNN&54.531&67.546&38.000&38.261&17.483&21.288&19.923&11.824\\
		ConvLSTM&81.155&115.292&61.783&24.590&26.054&20.643&23.479&11.110\\
		ST-ResNet&38.744&61.021&95.236&66.676&32.790&33.078&33.511&13.608\\
		DMVST-Net&26.504&27.418&37.083&79.421&40.295&26.635&38.955&51.070\\
		PCRN&26.879&21.868&20.620&40.321&14.811&26.793&13.038&\textbf{1.272}\\
		DeepSTN+&101.730&101.602&207.788&184.692&38.459&24.950&26.828&49.394\\
		STDN&43.319&56.437&16.428&\textbf{10.171}&\textbf{1.900}&11.223&7.506&18.041\\
		Multitask-DF&61.564&\textbf{3.396}&27.148&41.094&15.129&\textbf{0.007}&\textbf{6.681}&14.158\\
		VLUC-Nets (Pyramid)&\textbf{11.217}&45.914&\textbf{0.070}&37.548&3.782&8.936&7.699&2.650\\
		\hline
	\end{tabular*}
\end{table*}

\begin{table*}[h]
	\centering
	\caption{Effectiveness Evaluation of Crowd Density Prediction at Tokyo Disneyland (BousaiTYO Density, RMSE)}
	\label{tab:tokyodisneyland}
	\begin{tabular*}{15cm}{@{\extracolsep{\fill}}|l|c|c|c|c|c|c|c|c|}
		\hline
		\multicolumn{1}{|l|}{} &
		\multicolumn{4}{c|}{Weekday} &
		\multicolumn{4}{c|}{Weekend} 
		\\
		\hline
		\multicolumn{1}{|l|}{Model} & 
		\multicolumn{1}{c|}{08:00} & 
		\multicolumn{1}{c|}{12:00} &
		\multicolumn{1}{c|}{16:00} & 
		\multicolumn{1}{c|}{20:00} & 
		\multicolumn{1}{c|}{08:00} & 
		\multicolumn{1}{c|}{12:00} & 
		\multicolumn{1}{c|}{16:00} & 
		\multicolumn{1}{c|}{20:00} 
		\\
		\hline
		HistoricalAverage&46.500&19.000&17.786&26.643&43.133&18.733&6.600&7.667\\
		CopyYesterday&58.000&47.000&42.000&46.000&38.000&23.000&30.000&59.000\\
		CNN&9.447&12.679&5.577&7.747&5.868&\textbf{0.400}&2.330&11.803\\
		ConvLSTM&9.251&9.170&\textbf{0.462}&3.058&6.508&1.895&4.441&12.158\\
		ST-ResNet&13.126&9.128&4.291&12.289&\textbf{0.890}&2.912&0.727&\textbf{0.102}\\
		DMVST-Net&15.761&7.075&1.515&11.274&1.069&2.175&0.631&3.834\\
		PCRN&\textbf{7.769}&5.712&6.085&10.095&4.114&2.080&0.747&11.471\\
		DeepSTN+&25.266&15.173&10.018&11.436&14.299&10.398&12.125&15.217\\
		STDN&18.896&12.237&9.763&4.603&0.445&0.978&0.928&14.851\\		
		Multitask-DF&10.727&\textbf{5.557}&12.050&\textbf{2.570}&10.460&10.378&3.371&0.955\\
		VLUC-Nets (Pyramid)&11.443&11.543&8.268&8.278&3.647&1.790&\textbf{0.065}&4.736\\
		\hline
	\end{tabular*}
\end{table*}

\subsection{Performance Evaluation}
 
\subsubsection{Effectiveness Evaluation.}
The overall evaluation results on effectiveness are summarized in Table \ref{tab:bousaidata1}$\sim$\ref{tab:bousaidata2} for BousaiTYO-OSA, Table \ref{tab:opendata1}$\sim$\ref{tab:opendata2} for TaxiBJ, TaxiNYC, BikeNYC I-II. Through them we can see that the state-of-the-art models including our proposed  had advantages compared with baselines (HistoricalAverage$\sim$ConvLSTM). In particular, we have the following observations on newly published datasets and the existing ones.
\begin{itemize}
\item On bousai dataset, the state-of-the-arts (ST-ResNet$\sim$Multitask-DF) had their own advantages on different cities, tasks, and metrics, and none of them could achieve the best results from different points of view. The proposed VLUC-Nets with pyramid architecture became the dominant approach, and demonstrate the superiority to the state-of-the-arts as well as VLUC-Nets without pyramid architecture on all of datasets, tasks, and metrics. As the spatial domain (i.e., $H$,$W$=$80$,$80$) of the new published dataset is much larger than the existing datasets, the pyramid architecture plays a vital role in capture the spatial dependency. Next to VLUC-Nets (pyramid-version), STDN also achieved satisfactory performances on all axises. However, as the computation unit of STDN is pixel (mesh-grid), the big spatial domain will generate a huge amount of training samples, so the training process of STDN would take far more time than other models.

\item On the existing open datasets, STDN showed the best performances on TaxiBJ, BikeNYC-I, and BikeNYC-II in general, while VLUC-Nets (plain-version) achieved the best performance on TaxiNYC. Since the spatial domain of the existing datasets is relatively small, VLUC-Nets stacked with plain ConvLSTMs demonstrated better performance comparing with pyramid-version VLUC-Nets. Besides, on BikeNYC-I and BikeNYC-II, VLUC-Nets achieved the lowest MAE, and the performances on other three metrics were very close to STDN. On TaxiNYC, STDN could achieve the second best performance. On TaxiBJ, next to STDN, DeepSTN+ ranked at the second place, and VLUC-Nets ranked at the third place. In general, on four open datasets, STDN and plain-version VLUC-Nets hold relatively clear advantages over other models. Note that the training process of STDN was still the slowest one on any of the existing datasets. 
\end{itemize}	

In contrast to the evaluation from the overall view, we also conducted a very specific evaluation on 1 weekday (last Fri. in BousaiTYO, i.e. 2017/7/7) and 1 weekend (last Sat. in BousaiTYO, i.e. 2017/7/8) in two selected locations (mesh-grids) to show the ground-truth and predicted density. One is the mesh-grid containing Tokyo Station, a typical CBD area, another is the mesh-grid containing Tokyo Disneyland, a popular theme park. RMSE was calculated at four timestamps (08:00, 12:00, 16:00 and 20:00) on each day. From Table \ref{tab:tokyostation}$\sim$\ref{tab:tokyodisneyland}, we can see that the deep learning models had their own advantages at different types of day, times, and locations. In particular, VLUC-Nets had the lowest overall RMSE on BousaiTYO density in Table \ref{tab:bousaidata1}, but it did not overwhelm all the other models at the two case studies in Table \ref{tab:tokyostation}$\sim$\ref{tab:tokyodisneyland}. This demonstrates that the global optimal metric can't assure the local optimal in specific location and time.
How to achieve a global-local consistent model will be left as an open challenge and an independent topic that deserves further research effort.

\begin{table}[h]
	\centering
	\caption{Efficiency Evaluation on BousaiTYO Density and In-Out Flow}\label{tab:efficiency1}		
	\begin{tabular*}{15.5cm}{@{\extracolsep{\fill}}|l|c|c|c|c|c|c|}
		\hline
		\multicolumn{1}{|l|}{} &
		\multicolumn{3}{c|}{Tokyo Density} &
		\multicolumn{3}{c|}{Tokyo In-Out Flow} 
		\\
		\hline
		\multirow{2}{2cm}{Model} & Number of & Time (s) & Epochs & Number of & Time (s) & Epochs \\
		& Parameters & per Epoch & to Converge & Parameters & per Epoch & to Converge \\
		\hline
		CNN & 20,929 & 9 & 47 & 22,946 & 8 & 22 \\
		ConvLSTM & 187,432 & 108 & 80 & 189,848 & 107 & 94 \\
		ST-ResNet & 205,095 & 20 & 185 & 297,002 & 21 & 121\\
		DMVST-Net & 1,578,253 & 1,669 & 68 & 1,578,253 & 1,684 & 59\\
		PCRN & 1,284,999 & 300 & 169 & 1,561,352 & 300 & 167\\
		DeepSTN+ & 327,820,193 & 154 & 38 & 327,820,546 & 142 & 97\\
		STDN & 6,181,505 & 4,895 & 85 & 6,349,922 & 5100 & 43\\
		Multitask-DF & 266,176 & 150 & 166 & 266,176 & 150 & 166\\
		VLUC-Nets (Pyramid) & 7,882,842 & 356 & 107 & 8,315,355 & 515 & 138\\
		\hline
	\end{tabular*}
\end{table}

\begin{table}[h]
	\centering
	\caption{Efficiency Evaluation on BousaiOSA Density and In-Out Flow}\label{tab:efficiency2}	
	\begin{tabular*}{15.5cm}{@{\extracolsep{\fill}}|l|c|c|c|c|c|c|}
		\hline
		\multicolumn{1}{|l|}{} &
		\multicolumn{3}{c|}{Osaka Density} &
		\multicolumn{3}{c|}{Osaka In-Out Flow} 
		\\
		\hline
		\multirow{2}{2cm}{Model} & Number of & Time (s) & Epochs & Number of & Time (s) & Epochs \\
		& Parameters & per Epoch & to Converge & Parameters & per Epoch & to Converge \\
		\hline
		CNN & 20,929 & 7 & 44 & 22,946 & 7 & 16\\
		ConvLSTM & 187,432 & 79 & 68 & 189,848 & 79 & 119 \\
		ST-ResNet & 165,895 & 20 & 184 & 218,602 & 20 & 120\\
		DMVST-Net & 1,578,253 & 980 & 50 & 1,578,253 & 989 & 62\\
		PCRN & 806,199 & 170 & 166 & 962,152 & 155 & 192\\
		DeepSTN+ & 184,414,913 & 90 & 39 & 184,415,138 & 90 & 49\\
		STDN & 6,181,505 & 2,891 & 85 & 6,349,922 & 3,251 & 55\\
		Multitask-DF & 266,176 & 119 & 183 & 266,176 & 119 & 183\\
		VLUC-Nets (Pyramid) & 6,264,442 & 267 & 90 & 6,512,155 & 272 & 66\\
		\hline
	\end{tabular*}
\end{table}

\begin{table}[h]
	\centering
	\caption{Efficiency Evaluation on TaxiBJ and TaxiNYC}\label{tab:efficiency3}	
	\begin{tabular*}{15.5cm}{@{\extracolsep{\fill}}|l|c|c|c|c|c|c|}
		\hline
		\multicolumn{1}{|l|}{} &
		\multicolumn{3}{c|}{TaxiBJ} &
		\multicolumn{3}{c|}{TaxiNYC} 
		\\
		\hline
		\multirow{2}{2cm}{Model} & Number of & Time (s) & Epochs & Number of & Time (s) & Epochs \\
		& Parameters & per Epoch & to Converge & Parameters & per Epoch & to Converge \\
		\hline
		CNN & 22,946 & 28 & 21 & 22,946 & 4 & 67\\
		ConvLSTM & 189,848 & 333 & 81 & 189,848 & 45 & 74\\
		ST-ResNet & 145,984 & 85 & 61 & 123,402 & 11 & 75\\
		DMVST-Net & 1,578,253 & 1,222 & 46 & 1,578,253 & 62 & 83\\
		PCRN & 410,398 & 660 & 50 & 234,552 & 83 & 80\\
		DeepSTN+ & 33,607,666 & 161 & 33 & 20,519,858 & 13 & 80\\
		STDN & 6,349,922 & 5,950 & 42 & 6,349,922 & 181 & 34\\		
		VLUC-Nets (Plain) & 9,567,089 & 850 & 32 & 9,036,923 & 98 & 139\\
		\hline
	\end{tabular*}
\end{table}

\begin{table}[h]
	\centering
	\caption{Efficiency Evaluation on BikeNYC-I and BikeNYC-II}\label{tab:efficiency4}	
	\begin{tabular*}{15.5cm}{@{\extracolsep{\fill}}|l|c|c|c|c|c|c|}
		\hline
		\multicolumn{1}{|l|}{} &
		\multicolumn{3}{c|}{BikeNYC-I} &
		\multicolumn{3}{c|}{BikeNYC-II} 
		\\
		\hline
		\multirow{2}{2cm}{Model} & Number of & Time (s) & Epochs & Number of & Time (s) & Epochs \\
		& Parameters & per Epoch & to Converge & Parameters & per Epoch & to Converge \\
		\hline
		CNN & 22,946 & 6 & 57 & 22,946 & 4 & 60\\
		ConvLSTM & 189,848 & 74 & 95 & 189,848 & 47 & 75\\
		ST-ResNet & 124,618 & 18 & 54 & 123,402 & 11 & 72\\
		DMVST-Net & 1,529,101 & 107 & 83 & 1,578,253 & 64 & 68\\
		PCRN & 245,440 & 146 & 116 & 234,552 & 83 & 85\\
		DeepSTN+ & 32,554,482 & 32 & 62 & 20,519,858 & 12 & 64\\
		STDN & 6,349,922 & 355 & 45 & 6,349,922 & 181 & 37\\		
		VLUC-Nets (Plain) & 9,070,171 & 158 & 82 & 9,036,923 & 104 & 54\\
		\hline
	\end{tabular*}
\end{table}

\subsubsection{Efficiency Evaluation.} 
Besides the comparison in terms of prediction accuracy, we also provide an efficiency comparison in terms of computational time and neural network complexity between the different approaches, as these can play an important role in practice when deciding which approach to use in practice. Specifically, we study the total number of parameters, training time per epoch in second, and epochs needed to converge for each model on each dataset, which have been listed as Table \ref{tab:efficiency1}$\sim$\ref{tab:efficiency4}. Through the tables, we can see that: (1) the overall efficiency of our proposed VLUC-Nets especially the plain version was controlled at an acceptable level; (2) the training time of DMVST-Net and STDN were far more than others as they took mesh-grid as the computation unit; (3) the parameter number of DeepSTN+ was far more than others as it utilized fully-connected layer to capture the citywide spatial dependency; (4) ST-ResNet holds a very clear advantage over other models from the view of efficiency.  

\subsubsection{Summary.}
In summary, the state-of-the-art models designed different techniques to capture spatiotemporal dependency, however, none of them could be acknowledged as a dominant model at the current stage. Their main limitations are as follows: (1) ST-ResNet converts the video-like data to high-dimensional image-like data and uses a simple fusion-mechanism to handle different types of temporal dependency; (2) through the experiment, it was found PCRN took more epochs to converge and tended to cause overfitting; (3) DMVST-Net and STDN use local CNN to take grid (pixel) as computation unit, resulting in long training time (nearly 1 week) on Bousai dataset; (4) DeepSTN+ utilized a fully-connected layer in $ConvPlus$ block, which would result in a huge number of parameters in Tokyo area (over 0.3 billion); (5) Multitask-DF needs both density and in-out flow data for computing. Finally, we can do the following recommendations based on the different requests and situations: (1) if effectiveness comes first, pyramid VLUC-Nets will be recommended for datasets with large spatial domain, and STDN will be recommended for small-spatial-domain data; (2) if efficiency comes first, ST-ResNet could be a very light but still powerful solution; (3) if we balance the effectiveness and efficient, plain-version VLUC-Nets could always be a good choice.

\section{Discussion}
\noindent\textbf{Challenges.} Although we try our best to make the benchmark as thorough as possible, there are still some open challenges left to be solved or to be further enhanced.
\begin{itemize}
	\item Multi-step predictability. In our benchmark, crowd/traffic density and in-out flow are defined as a single-step prediction problem.
	The multi-step predictability could be further validated, and an effective multi-step VLUC-Net is also considered to be proposed.
	\item Better explainability. Currently, the-state-of-the-arts and the proposed VLUC-Nets still lack of transparent explainability, which has been a major drawback of deep learning-based approaches. As a specific urban computing scenario, we hope to add more explainability to the models using the domain knowledge of geoinformation science.
	\item Thorough comparison. The current benchmark integrates the most relevant deep learning models following the similar problem definition. However, some trajectory-based prediction models could also be implemented as comparison models such as CityMomentum\cite{fan2015citymomentum} and DeepMove\cite{feng2018deepmove}. Moreover, some pure video prediction models like PredNet\cite{lotter2016deep} proposed and succeeded in the filed of computer vision could also be employed as baseline models.
\end{itemize}

\noindent\textbf{Applications.} This work will be a highly-extendable benchmark, which could be easily applied to other urban computing problems (other than citywide crowd/traffic prediction problems) as listed below. We would like to explore the extendability to these urban computing scenarios in the future.
\begin{itemize}
	\item Single-channel urban video. Citywide air quality can be represented with an urban tensor ($Timestep$, $Height$, $Width$, $Channel$=1), where $Channel$ stores the PM2.5 value for each mesh-grid. Similarly, citywide electric power consumption could also be represented with that tenor, where $Channel$ stores the aggregated data collected from each electric power meter of the houses inside the mesh-grid.
	\item Multi-channel urban video. People in city have different transportation behaviors such as by car, train/subway, bike or walk. Citywide transportation demand can be represented by ($Timestep$, $Height$, $Width$, $Channel$=4), where each $Channel$ corresponds to the number of people with one certain transportation mode. Similarly, citywide emergency incidents can also be represented as a multi-channel urban video, where each $Channle$ corresponds to a specific type of incident, such as crime, medical, and fire incidents. 
	\item Super-channel urban video. Inflow and outflow can only indicate how many people will flow into or out from a certain mesh-grid, and can't answer where (which mesh-grid) the people flow come or transit. For a crowd of inside one mesh-grid, their next locations could distributed over the entire mesh $\Omega$=($Height$,$Width$). Therefor, citywide detailed crowd-flow could be represented with a tensor ($Timestep$, $Height$, $Width$, $Channel$=$\Omega$), where $Channel$ is super-high dimension. 
\end{itemize}

\section{Related Work}
In this section, we briefly discuss some existing researches concerning to crowd or traffic prediction problem in the field of urban computing\cite{zheng2014urban}. Except for the video-like deep learning approaches, there are also some statistical models proposed. Spatiality preservable factored Poisson regression \cite{shimosaka2019spatiality} incorporates point of interest to overcome data sparsity and degraded performance in even finer grains population prediction. CityProphet\cite{konishi2016cityprophet,shimosaka2015forecasting} and \cite{zhou2018early} utilize query data of Smartphone App to forecast only crowd density other than crowd flow. \cite{akagi2018fast,sudo2016particle} conduct transition estimation from aggregated population data, and \cite{tanaka2018estimating} estimates the transition populations using inflow and outflow defined by \cite{hoang2016forecasting}. 

Based on road network, \cite{castro2012urban,huang2014deep,abadi2014traffic,lv2015traffic,ma2015large,ma2015long,song2016deeptransport} were proposed to predict the traffic flow, speed, congestion, human mobility as well as transportation mode. In particular, DeepTTE\cite{wang2018learning} and DeepGTT\cite{li2019learning} are proven as effective deep learning models to predict the travel time on each road segment. Leveraging on the latest techniques, a series of models have been proposed to address traffic-related problems, such as using graph neural networks for traffic forecasting \cite{yu2018spatio} and ride-hailing demand prediction \cite{geng2019spatiotemporal}, multitask learning for travel time estimation \cite{li2018multi}, or meta learning for traffic prediction \cite{pan2019urban}.

Besides, many trajectory-based deep learning models were proposed to predict each individual's movement \cite{zhang2016gmove,liu2016predicting,gao2019predicting}. \cite{liu2016predicting} extends a regular RNN by utilizing time and distance specific transition matrices to propose an ST-RNN model for predicting the next location. DeepMove \cite{feng2018deepmove}, considered to be a state-of-the-art model for trajectory prediction, designed a historical attention module to capture periodicities and augment prediction accuracy. VANext\cite{gao2019predicting} further enhanced DeepMove by proposing a novel variational attention mechanism. Modeling human mobility for very large populations \cite{song2010modelling,fan2015citymomentum} and simulating human emergency mobility following disasters \cite{song2013modeling,song2015} are similar problems to ours; however, their models are built based on millions of individuals' mobility. \cite{fan2014} propose tensor factorization approach to decompose urban human mobility, aiming to understand basic urban life patterns from city-scale human mobility data. Using mobility data from location-based social networks (LBSN), \cite{lian2014geomf,wang2015regularity} design matrix factorization-based models to conduct PoI recommendation or location prediction. Also, \cite{yin2017spatial} utilized latent factor models for POI recommendation using heterogeneous features. \cite{guo2018citytransfer} conducted chain store site recommendation with transfer learning and multi-source data. 

Last, other video-like urban computing problems are also modeled based on citywide mesh-grids, and addressed through advanced deep learning technologies, including air quality prediction\cite{lin2018exploiting,yi2018deep}, crop yield prediction\cite{you2017deep}, crime prediction\cite{huang2018deepcrime}, and abnormal event prediction\cite{huang2019mist}.

\section{Conclusion and Future Work}
In this study, we build a standard benchmark called VLUC for video-like computing on citywide traffic and crowd prediction by publishing new datasets called BousaiTYO and BousaiOSA and integrating the existing ones including TaxiBJ, BikeNYC I-II, and TaxiNYC. An online visualization tool called Mobmap can function like a typical video player on those data. We comprehensively and systematically review the state-of-the-art works of literature including ST-ResNet, DMVST-Net, PCRN, STDN, DeepSTN+, and Multitask-DF, and conduct a thorough performance evaluation for those models on density and in-out flow prediction problems. We empirically design a family of models named as VLUC-Nets by employing the advanced deep learning techniques to more effectively capture spatial and temporal dependency. The experimental results demonstrate the superior performances of VLUC-Nets to the state-of-the-arts on both new datasets and the existing ones. This benchmark \url{https://github.com/deepkashiwa20/VLUC} will be officially published including the new dataset if this paper is accepted. 

In the future, apart from the used metrics like RMSE, MAE, and MAPE, some supplemental metrics are considered to be introduced, such as Hausdorff Distance, Longest Common Subsequence (LCSS), and Dynamic Time Warping (DTW) to verify the time-series similarity, or KL-Divergence and Cross Entropy to verify probability distribution. Then, the effects of different settings on the use of heterogeneous data sources such as POI and event info, objective function, and scaling strategy could be further analyzed. Moreover, we will continue to explore the limitations of the state-of-the-arts as well as our proposed models, and try to improve the performances on the benchmark. Lastly, as discussed above, we will try to address the open challenges and apply this family of approaches to other video-like urban computing problems, such as citywide air quality prediction, and citywide crime incident prediction.

\section*{Acknowledgments}
\bibliographystyle{ACM-Reference-Format}
\bibliography{reference}

\end{document}